\documentclass[11pt]{article}

\usepackage[final]{acl}
\usepackage{multirow}
\usepackage{times}
\usepackage{float}
\usepackage{algorithm}
\usepackage{algorithmic}
\usepackage{latexsym}
\usepackage{amsmath}
\usepackage{booktabs}
\usepackage[T1]{fontenc}
\usepackage{amsmath}
\usepackage{amssymb}  

\usepackage[utf8]{inputenc}
\usepackage{xcolor}
\usepackage{multirow}
\usepackage{makecell}
\usepackage{microtype}

\usepackage{inconsolata}

\usepackage{graphicx}
\usepackage{xcolor}  
\usepackage{CJKutf8}
\usepackage{xparse}
\usepackage{enumitem} 

\usepackage{rotating}    

\title{Rethinking Retrieval-Augmented Generation as a Cooperative Decision-Making Problem}

\renewcommand\footnotemark{}   

\author{Lichang Song \and Ting Long$^*$ \and Yi Chang$^*$ \\
    Jilin University \\
    \texttt{songlc24@mails.jlu.edu.cn, longting@jlu.edu.cn, yichang@jlu.edu.cn} \\
    \thanks{$^*$ Corresponding author}
}

\begin{document}
\maketitle
\begin{abstract}

Retrieval-Augmented Generation (RAG) has demonstrated strong effectiveness in knowledge-intensive tasks by grounding language generation in external evidence. Despite its success, many existing RAG systems are built based on a ranking-centric, asymmetric  dependency paradigm, where the generation quality of the generator is highly dependent on reranking results of the reranker.
To overcome this limitation, we propose Cooperative Retrieval-Augmented Generation (CoRAG), a framework that treats the reranker and the generator as peer decision-makers rather than being connected through an asymmetric dependency pipeline. By jointly optimizing their behaviors toward a shared task objective, the reranker and generator are encouraged to cooperate, ensuring that document reranking and generation work in concert to improve the final response.
Experimental results demonstrate good generalization and improved generation stability of CoRAG, even when the model is trained on only around 10K PopQA samples. Our model released in \url{https://github.com/CoderrrSong/CoRAG}

\end{abstract}

\section{Introduction}
In recent years, Retrieval-Augmented Generation (RAG) \cite{lewis2020retrieval, asai2024self, gao2023retrieval, zhao2024retrieval} has emerged as an important paradigm for enhancing the factuality and knowledge coverage of large language models (LLM) \cite{gao2023retrieval}. A typical RAG system consists of two core components: a retriever and a generator \cite{lewis2020retrieval,oche2409systematic}. Given a query, the retriever retrieves candidate documents from a large external corpus and further rerank them via reranker.
The generator then conditions on the query and the reranked documents to produce the final response. By explicitly incorporating external knowledge during generation, RAG effectively mitigates hallucinations and achieves strong performance on open-domain question answering tasks \cite{lewis2020retrieval,asai2024self}.
\begin{figure}[t]
    \centering
    \includegraphics[width=\columnwidth]{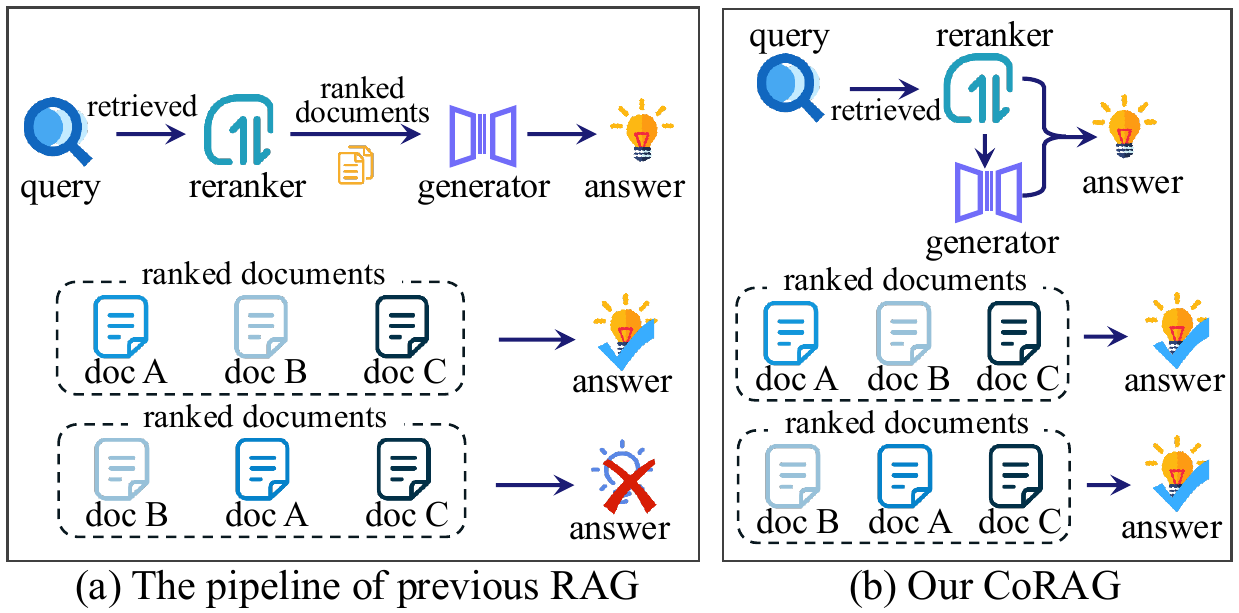}  
    \caption{
       Comparison with previous works.
    Previous works assume an asymmetric  dependency between the reranker and the generator, 
    whereas CoRAG treats them as equal participants optimized under a shared task-oriented reward.
    }
    \label{fig:intro} 
\end{figure}


As the reranker plays a critical role in shaping the document context provided to the generator \cite{lewis2020retrieval,sharma2025retrieval}, a growing body of recent work has focused on the design and optimization of rerankers and generators \cite{gao2023retrieval,asai2024self,sun2025dynamicrag,shen2023joint,jia2025bridging}.
However, most existing RAG methods still adopt a ranking-centric, asymmetric-dependency paradigm (Figure 1(a)), where the reranker produces a fixed document ordering and the generator performs generation conditioned on these top-ranked documents. This design tightly couples generation with reranking decisions, making the generator highly sensitive to the reranking results.
As shown in Figure 1(a), if a suboptimal reranker misranks a less relevant document at the top, the generator may produce an incorrect response, even though the optimal document is still present within the top-N set.

From an optimization perspective, this phenomenon reveals a deeper mismatch between reranking and generation. On the one hand, the generator is highly sensitive to fine-grained ranking results; on the other hand, 
learning an exact total order over multiple highly relevant documents is inherently harder than learning a relaxed ordering for the reranker.
For instance, precisely ranking the top three relevant documents as positions 1, 2, and 3 is substantially more challenging than merely ensuring that these documents appear within the top few positions in any order. 

A natural way to address this issue is to treat the reranker and generator as cooperative decision-makers optimizing a shared reward (Figure \ref{fig:intro}(b)). 
The cooperative multi-agent reinforcement learning (MARL) framework \cite{chen2025improving} provides a natural formulation for this problem,  where the reranker and generator can coordinate and adapt to each other's needs. By jointly optimizing under a shared reward, the reranker is encouraged to learn a more accurate total order targeting the final results while relaxing the generator's asymmetric dependency on reranking quality.
However, realizing this idea with standard MARL optimizers is challenging. A fundamental issue is that MARL's underlying MDP assumption structurally mismatches the core properties of ranking tasks. For example, MARL agents may independently select the same document for different positions, leading to invalid or degenerate rankings, which further compromises downstream generation \cite{chen2025improving}.

In this paper, we propose a method called Cooperative Retrieval-Augmented Generation (CoRAG), which models the retrieval-augmented generation as a cooperative decision-making problem and devises a novel GRPO-style optimization for the reranker within the MARL framework.
Unlike standard MARL optimizers that suffer from the MDP-ranking mismatch, our GRPO-style design makes the otherwise challenging MARL-based reranker optimization tractable.
As a result, CoRAG mitigates the generator’s asymmetric dependency to fine-grained reranking results and improves generation stability.
Experimental results show that, despite being trained on only around 10K PopQA samples, CoRAG significantly outperforms baselines and generalizes well across multiple datasets and tasks.

In summary, our contributions are:

\begin{itemize} [left=-5pt]
\item 
We identify and formalize the asymmetric dependency between the reranker and generator in RAG, and propose a joint MARL optimization scheme to mitigate it.
\item We propose a GRPO-style  optimization for the reranker within a MARL framework to overcome the MDP-ranking mismatch.
\item Extensive experiments demonstrate improved robustness and generalization of our CoRAG.
\end{itemize}




\section{Problem Definition}

A typical RAG system consists of two components: a retriever and a generator \cite{lewis2020retrieval,oche2409systematic}.
Given an input query, the retriever retrieves a candidate document set from a large external corpus, which is further refined by a reranker before being used by the generator to generate the final response.
In this work, we focus on the reranker in the retrieval stage together with the generator, as reranking plays a crucial role in improving the relevance and faithfulness of generation \cite{sharma2025retrieval}.
We model RAG as a cooperative multi-agent decision-making problem, treating the reranker and the generator as two cooperative agents that jointly determine the final generation outcome, and train them with the goal of maximizing a shared task-oriented objective defined on the generated response.


Specifically, given a query $q$ and a candidate document set $\mathcal{D}$, 
the reranker $\mathcal{S}_\theta$ reranks and selects a set of documents $D \subseteq \mathcal{D}$,
and the generator $\mathcal{G}_\phi$ generate a response $\hat{a}$ conditioned on $q$ and $D$.
The learning objective is defined as
\begin{equation}
\max_{\theta, \phi} \;
\mathbb{E}_{D \sim \mathcal{S}_\theta(\cdot \mid q, \mathcal{D}), \,
\hat{a} \sim \mathcal{G}_\phi(\cdot \mid q, D)}
\left[ R(a^{\ast}, \hat{a}) \right],
\end{equation}
where $\theta$ and $\phi$ are the parameters of the reranker and generator respectively. $R(a^{\ast}, \hat{a})$ denotes a task-oriented reward defined on the generated response.
This formulation emphasizes that the reranker is optimized not for ranking accuracy, but for its contribution to downstream generation quality. Similarly, the generator is trained to produce outputs that effectively utilize the provided document configuration to maximize the same task-oriented reward.
By aligning both components to the same outcome-oriented objective, the reranker and generator are encouraged to cooperate, 
ensuring that document reranking and generation in concert to improve the final response.

\section{Method}
\begin{figure}[t]
    \centering
    \includegraphics[width=0.45\textwidth]{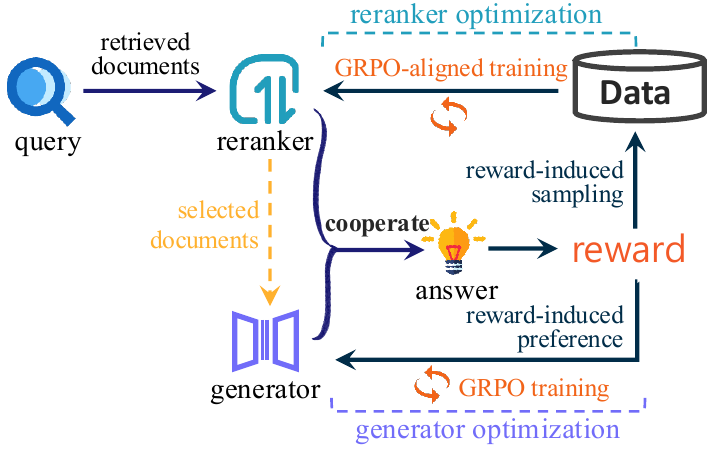}  
    \caption{CoRAG overview.
The reranker and generator cooperate to generate responses. The task-oriented reward derived from response guides GRPO-aligned training of the reranker and GRPO optimization of the generator.}
    \label{fig:method} 
\end{figure}

An overview of our proposed Cooperative Retrieval-Augmented Generation (CoRAG) is shown in Figure \ref{fig:method}.
In the following sections, we first describe the reranker and generator in our CoRAG, and then discuss their joint optimization.


\subsection{The Reranker}

The reranker aims to refine the retrieved candidate document set $\mathcal{D}$ by re-ranking the documents according to their relevance, and selects a subset $D$ to provide to the downstream generator for response generation.
Specifically, we obtain $D$ in the following steps:

Given a query $q$ and a document $d_i$ in the candidate document set $\mathcal{D} = \{d_1, d_2, ..., d_N\}$, we compute the relevance score of document $d_i$ to $q$: 
\begin{equation}
s_i 
= \mathcal{S}_{\theta}(q, d_i),
\end{equation}
$\mathcal{S}_{\theta}$ denotes the reranker and we implemented with BGE-Reranker \citep{multi2024m3}. 

We select a subset of top-$K$ documents according to the relevance scores and feed them into the generator.
\begin{equation} \label{eq:topk}
D = \left\{ d_i \;\middle|\; i \in \operatorname{Top\text{-}K}(\{s_1,\dots,s_N\}) \right\}
\end{equation}




\subsection{The Generator}

The Generator is responsible for generating the response $\hat{a}$ based on the selected document $D$ and the query $q$. Specifically, we obtain the predicted response $\hat{a}$ with the following steps:

Given a query $q$ and the top-K documents $D$, we input them into the generator model $\mathcal{G}_{\phi}$ to generate the final response:
\begin{equation}
    \hat{a} = \mathcal{G}_{\phi}(q, D),
\end{equation}
where $\mathcal{G}_{\phi}$ is the generator, typically implemented as an autoregressive language model \cite{radford2018improving, brown2020language} that generates the response token by token conditioned on both the query $q$ and the document $D$.

\subsection{The Optimization}
The reranker and the generator are optimized under a shared task-oriented reward.
From a multi-agent perspective, we treat the reranker and the generator as two cooperative agents with distinct decision roles. The reranker determines which documents to attend to, while the generator decides how to synthesize the final response based on the selected documents. The shared reward, defined on the final response, couples their behaviors and enables coordinated optimization toward task-oriented success.

\begin{equation}
r = \mathrm{R}(a^\ast, \hat{a}),
\end{equation}
where $\mathrm{R}(a^\ast, \hat{a})$ denotes a task-oriented evaluation function, which we implement to return 1 if the generated response $\hat{a}$ contains the ground-truth response $a^\ast$, and 0 otherwise. 
In the following, we detail the optimization of the reranker and the generator under this shared task-oriented reward.


\paragraph{Reranker Optimization.}
Unlike standard learning-to-rank settings \cite{casalegno2022learning}, document-level supervision is not directly available.
To address that, we transform task-oriented rewards into document-level stochastic preference signals for reranker optimization. Since these signals indicate how documents collectively contribute to task success, we optimize the reranker in a group-relative preference aligned with GRPO \cite{shao2024deepseekmath}.

First, to transform delayed task-level rewards into document-level stochastic preference signal,  
for each document $d_i$  in the top-K document set $D$ at training iteration $t$, we assign a binary success signal $l^{(t)}(q, d_i) \in \{0,1\}$, indicating whether the generated response achieved task success when $d_i$ was included.
Although this signal provides only a coarse and noisy approximation of individual document contribution under multi-document conditioning, it enables scalable credit assignment without requiring explicit supervision.



Based on these signals,we estimate the expected task success associated with each document $d_i$:

\begin{equation}
\bar{l}(q,d_i) = \frac{1}{T}\sum_{t=1}^{T} l^{(t)}(q,d_i),
\end{equation}
where $T$ denotes the number of historical iterations. To account for uncertainty, we map $\bar{l}(q,d_i)$ to a smoothed Bernoulli parameter
\begin{equation}
p(q,d_i) = \alpha + (1 - 2\alpha)\cdot \bar{l}(q,d_i),
\end{equation}
where $\alpha \in (0, 0.5)$, and we sample a stochastic preference label by:
\begin{equation}
p_i \sim \text{Bernoulli} \left(p(q,d_i) \right),
\quad  p_i \in \{0,1\}.
\end{equation}
This  stochastic preference label serves as a surrogate feedback signal for reranker optimization,
enabling us to optimize the reranker in a group-relative preference framework aligned with GRPO. 
To conduct the optimization,
we model the reranker as a deterministic scoring function $\mathcal{S}_\theta(q,d_i)$ that induces a distribution over candidate documents: $\pi_\theta(d_i \mid q, \mathcal{D}) \propto \exp(\mathcal{S}_\theta(q,d_i))$.
The group-relative advantage $\hat{A}(q,d)$ is derived from these preference labels: positive labels ($p_i=1$) indicate relative advantage  for document $d_i$ within the group, while a negative label indicates a disadvantage. This leads to the GRPO objective:
\begin{equation}
\mathcal{L}_{\mathrm{GRPO}}^{r}
=
- \mathbb{E}_{d_i \sim \pi_\theta}
\left[
\hat{A}(q,d_i)\log \pi_\theta(d_i \mid q, \mathcal{D})
\right],
\end{equation}

However, directly optimizing the stochastic objective suffers from high variance under noisy credit estimates.
We therefore reinterpret the group-relative advantages as inducing pairwise preferences among candidate documents: documents with higher empirical success rates (more likely to receive positive labels) should be ranked higher. This allows us to reduce GRPO to a deterministic learning-to-rank problem. Constructing positive and negative sets $\mathcal{D}^+$ and $\mathcal{D}^-$ from the sampled labels, we adopt a margin-based pairwise ranking surrogate: 
\begin{equation} \label{ea:rank_loss}
\small
\mathcal{L}_{\mathrm{rank}}
=
\sum_{d_i^+ \in \mathcal{D}^+}
\sum_{d_j^- \in \mathcal{D}^-}
\max\!\left(0,\;
s_\theta(q,d_j^-) - s_\theta(q,d_i^+) + \gamma
\right), 
\end{equation}
where $\gamma$ is the margin hyperparameter. Although this reduction is not a strict equivalence to GRPO, the ranking loss preserves the group-relative preferences induced by the stochastic labeling scheme and the underlying GRPO objective in expectation. It thus provides a stable and efficient surrogate for optimizing the reranker while maintaining alignment with task-oriented rewards.


\paragraph{Generator Optimization.}
The generator defines a conditional generation policy
$\pi_\phi(\hat{a} \mid q, D)$, where $D$ is the set of top-K documents selected
by the reranker.
The generator is optimized using standard GRPO for conditional text
generation:
\begin{equation}
\mathcal{L}_{\text{gen}}
=
- \mathbb{E}_{\hat{a} \sim \pi_\phi}
\left[
\hat{A}(\hat{a})
\log \pi_\phi(\hat{a} \mid q, D)
\right],
\end{equation}
where the group-relative advantage $\hat{A}(\hat{a})$ is computed from the
task-oriented reward $r$ using a baseline that compares the current response to other responses in the same training batch. 

Overall, both the reranker and the generator are optimized with respect to
the same task-oriented reward, enabling them to cooperatively improve retrieval relevance and generation quality.

\section{Experiments}
We conduct extensive experiments to evaluate the performance of CoRAG, and we particularly focus
on the research questions: (i) Is CoRAG more effective than existing methods \textbf{(RQ1)}? 
(ii) How important are the reranker and the generator to overall performance, and is it necessary to jointly optimize them \textbf{(RQ2)}?
(iii) How does performance vary with the number of documents used \textbf{(RQ3)}? (iv) Does CoRAG generalize to other tasks \textbf{(RQ4)}? (v) Does the CoRAG generator produce high-quality outputs as judged by other LLMs \textbf{(RQ5)}?



\subsection{Experimental Setting}
\paragraph{Datasets}
We evaluate our method on multiple knowledge-intensive benchmarks, following the dataset setup of recent works \cite{wei2024instructrag}. Specifically, we evaluate on PopQA \cite{mallen2023not}, TriviaQA \cite{joshi2017triviaqa}, Natural Questions (NQ) \cite{kwiatkowski2019natural}, ASQA \cite{stelmakh2022asqa}. Additionally, we include 2WikiMultiHopQA \cite{ho2020constructing}, a Wikipedia-based cross-document multi-hop QA dataset that requires models to reason across multiple documents to derive responses.

\begin{table}[h]
\centering
\caption{Dataset statistics.}
\resizebox{0.75\linewidth}{!}{%
\small
\begin{tabular}{lccc}
\hline
Dataset & Train  & Test   \\
\hline
PopQA & 12,868  & 1,399 \\
TriviaQA & 78,785 & 11,313 \\
NaturalQuestions & 79,168 & 3,610   \\
ASQA & 4,353& 948 \\
2WikiMultiHopQA & 167,454 & 12,576 \\
\hline
\end{tabular}%
}
\label{tab:dataset_stats}
\end{table}

\paragraph{Evaluation Metrics.} 
Following InstructRAG \cite{wei2024instructrag}, 
we report correctness, citation precision, and citation recall \cite{gao2023enabling} for ASQA. For the other datasets, we use accuracy to measure whether the ground-truth responses are included in the generated outputs.

\begin{table*}[t]
\centering
\caption{Overall results of our method and baselines on five benchmarks. Baseline results are reported in InstructRAG \cite{wei2024instructrag}. - indicates the results are not reported or not applicable.
The best and second-best performances are highlighted in bold and with an underline, respectively. Notably, our model is trained only on PopQA \cite{mallen2023not}, while all other datasets are used exclusively for evaluation.}

\small
\begin{tabular}{lccccccc}
\hline
\multirow{2}{*}{Method} & PopQA & TriviaQA & NQ & 2WikiMultiHopQA & \multicolumn{3}{c}{ASQA} \\
\cline{6-8}
 & (acc) & (acc) & (acc) & (acc) & (em) & (pre) & (rec) \\
\hline
\multicolumn{8}{c}{\textit{Baselines w/o Retrieval}} \\
\hline
\textbf{Vanilla Zero-shot Prompting} & & & & & & & \\
\quad ChatGPT & 29.3 & 74.3 & -- & -- & 35.3 & -- & -- \\
\quad Llama-3-Instruct$_{8\text{B}}$ & 22.8 & 69.4 & 46.6 & 45.6 & 30.6 & -- & -- \\
\quad Llama-3-Instruct$_{70\text{B}}$ & 28.9 & 80.6 & 57.9 & 57.5 & 39.1 & -- & -- \\
\hline
\multicolumn{8}{c}{\textit{RAG w/o Training}} \\
\hline
\textbf{In-Context RALM} \cite{ram2023context} & & & & & & & \\
\quad ChatGPT & 50.8 & 65.7 & -- & -- & 40.7 & 65.1 & 76.6 \\
\quad Llama-3-Instruct$_{8\text{B}}$ & 62.3 & 71.4 & 56.8 & 43.4 & 40.0 & 62.1 & 66.4 \\
\quad Llama-3-Instruct$_{70\text{B}}$ & 63.8 & 76.3 & 60.2 & 51.2 & 43.1 & 62.9 & 67.6 \\
\textbf{Few-Shot Demo. w/ Instruction} & & & & & & & \\
\quad Llama-3-Instruct$_{8\text{B}}$ & 63.1 & 74.2 & 60.1 & 45.3 & 42.6 & 55.0 & 64.4 \\
\quad Llama-3-Instruct$_{70\text{B}}$ & 63.9 & 79.1 & 62.9 & 53.9 & 45.4 & 49.3 & 57.1 \\
\textbf{InstructRAG-ICL} \cite{wei2024instructrag}& & & & & & & \\
\quad Llama-3-Instruct$_{8\text{B}}$ & 64.2 & 76.8 & 62.1 & 50.4 & 44.7 & 70.9 & 74.1 \\
\quad Llama-3-Instruct$_{70\text{B}}$ & 65.5 & 81.2 & 66.5 & 57.3 & 47.8 & 69.1 & 71.2 \\

\hline
\multicolumn{8}{c}{\textit{RAG w/ Training}} \\
\hline
\textbf{Vanilla Supervised Fine-tuning} & & & & & & & \\
\quad Llama-3-Instruct$_{8\text{B}}$ & 61.0 & 73.9 & 56.6 & 56.1 & 43.8 & -- & -- \\
\textbf{Self-RAG} \cite{asai2024self} & & & & & & & \\
\quad Llama-2$_{7\text{B}}$ & 55.8 & 68.9 & 42.4 & 35.9 & 30.0 & 66.9 & 67.8 \\
\quad Llama-2$_{13\text{B}}$ & 56.3 & 70.4 & 46.4 & 36.0 & 31.4 & \textbf{70.3} & \textbf{71.3} \\
\quad Llama-3-Instruct$_{8\text{B}}$ & 55.8 & 71.4 & 42.8 & 32.9 & 36.9 & \underline{69.7} & 69.7 \\
\textbf{RetRobust} \cite{yoran2023making} & & & & & & & \\
\quad Llama-2$_{13\text{B}}$ & -- & -- & 39.6 & 51.5 & -- & -- & -- \\
\quad Llama-3-Instruct$_{8\text{B}}$ & 56.5 & 71.5 & 54.2 & 54.7 & 40.5 & -- & -- \\
\textbf{InstructRAG-FT} \cite{wei2024instructrag} & & & & & & & \\
\quad Llama-3-Instruct$_{8\text{B}}$ & \underline{66.2} & \underline{78.5} & \underline{65.7} & \underline{57.2} & \textbf{47.6} & 65.7 & \underline{70.5} \\
\textbf{CoRAG} & & & & & & & \\
\quad Llama-3-Instruct$_{8\text{B}}$  & \textbf{71.2} & \textbf{81.0} & \textbf{72.4} & \textbf{58.2} & \underline{45.8} & 54.9 & 48.9 \\
\hline
\end{tabular}
\label{tab:results}
\end{table*}

\paragraph{Baselines.} Following InstructRAG \cite{wei2024instructrag}, 
we compare our CoRAG with three groups of method: Baselines without Retrieval, relying solely on parametric knowledge, including ChatGPT, Llama-3-Instruct\(_{8\text{B}}\), and Llama-3-Instruct\(_{70\text{B}}\); RAG without Training, leveraging retrieved documents via in-context learning or prompting, such as In-Context RALM \cite{ram2023context}, Few-shot Demonstration with Instruction, and InstructRAG-ICL \cite{wei2024instructrag}; and RAG with Training, involving explicit training in the retrieval-augmented framework, including Self-RAG \cite{asai2024self} for iterative improvement via self-retrieval, RetRobust \cite{yoran2023making} for noise robustness training, and InstructRAG-FT \cite{wei2024instructrag} for instruction-following fine-tuning with retrieval.

\paragraph{Implementation Details}

We adopt BGE-reranker-v2-m3 \cite{multi2024m3} as the reranker and Llama-3-Instruct\(_{8\text{B}}\)
\cite{dubey2024llama} as the generator. During training, 
we use Llama-3-Instruct\(_{8\text{B}}\) to provide coarse annotations for positive and negative documents in the training dataset, which helps the reranker learn more effectively by alleviating the sparsity of stochastic preference labels at the early stages of training.
We set the learning rate of reranker to 5e-5, and the learning rate of generator to 1e-5. Both reranker and generator adopt the LoRA fine-tuning strategy for efficient parameter updating. We set $\gamma = 1$ in Eq.\eqref{ea:rank_loss}, and the top-K selected documents $K=1$ to precisely attribute the impact of individual documents on task success. During inference, we set $K=3$ for PopQA, TriviaQA, NQ and ASQA, and $K=7$ for 2WikiMultiHopQA.
The generator uses a temperature coefficient of 0.7 to balance generation diversity and stability.
It is worth noting that our CoRAG is trained only on PopQA \cite{mallen2023not}, while all other datasets are used solely for evaluation, both due to our limited computational resources and to enable assessment of the generalization ability of our CoRAG.

\subsection{Main Results (RQ1)}
The results are reported in Table \ref{tab:results}. As it is shown, our method (CoRAG) achieves outstanding performance.
Notably, unlike the InstructRAG
series of methods that perform separate training on each dataset, our model is trained only on PopQA.
Even so, our method yields state-of-the-art results on four core tasks (PopQA, TriviaQA, NQ and 2WikiMultiHopQA), with the corresponding accuracy rates reaching 71.2\%, 81.0\%, 72.4\% and 58.2\% respectively, which significantly outperform existing methods such as RetRobust and  InstructRAG-FT. This can be attributed to the joint optimization framework, in which the reranker progressively selects more relevant documents and the generator continuously improves its ability to extract and integrate information from them.


However, our CoRAG underperforms on the ASQA dataset. We attribute this primarily to the task discrepancy: ASQA requires synthesizing answers from multiple sources and handling ambiguous questions, which differs substantially from the factoid-style, single-answer questions prevalent in our training data (PopQA). Consequently, the retrieval-generator synergy optimized on factoid QA does not generalize effectively to this more complex, multi-answer setting.

Beyond overall performance gains, our method shows strong robustness to reranking order. Baselines degrade significantly with shuffled documents, while CoRAG remains stable (see Appendix~\ref{appendix:C}, Table~\ref{tab:shuffle} for detailed experiments).

\subsection{Ablation Study (RQ2)}
\label{sec:ab}
To investigate the importance of the reranker and the generator to overall performance, as well as whether it is necessary to jointly optimize them, we conduct ablation studies.
Specifically, we have five variants:
\begin{itemize} [left=-5pt]
\item  \textbf{RTrain}: only finetune the reranker with the labels annotated by Llama-3-Instruct\(_{8\text{B}}\).
\item  \textbf{GTrain}: only finetune the generator with GRPO.
\item  \textbf{RGReplace}: Replace the generator of CoRAG with Llama-3-Instruct\(_{8\text{B}}\), and replace the reranker with BGE-Reranker in inference.
\item  \textbf{GReplace}: Replace the generator of CoRAG with Llama-3-Instruct\(_{8\text{B}}\) in inference.
\item  \textbf{RReplace}: Replace the reranker of CoRAG with BGE-Reranker in inference.

\end{itemize}
The result presented in Table \ref{tab:ablation}
, we can observe that: 
(1) The individually trained RTrain and GTrain underperform CoRAG across all four datasets, which implies that a multi-agent cooperative optimization formulation may better capture the interaction between the reranker and the generator, leading to improved performance compared to independent optimization.
(2) Comparing RGReplace, GReplace and CoRAG reveals that the generator contributes more significantly to performance improvement, whereas the reranker plays a relatively limited role. We attribute this phenomenon to the joint optimization. Because the reranker and generator are jointly optimized, the generator may achieve strong performance even under suboptimal reranking signals, potentially weakening the learning pressure on the reranker and reflecting a trade-off in the current design.

\begin{figure}[t]
    \centering
    \includegraphics[width=0.5\textwidth]{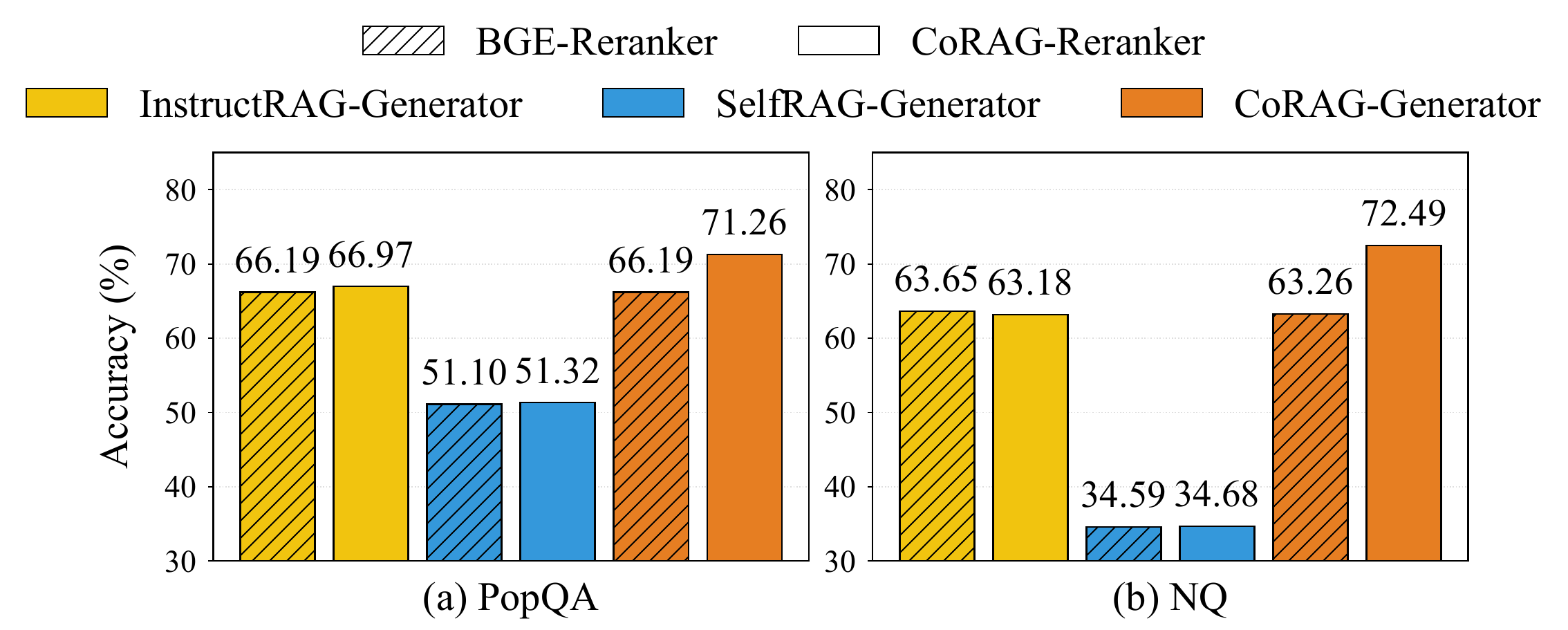}  
    \caption{Cross-validation results with different reranker and generator combinations (Top-3 setting). } 
    \label{fig:cross} 
\end{figure}

To examine this hypothesis, we conduct cross-component experiments by swapping rerankers and generators between CoRAG and other RAG frameworks (Self-RAG and InstructRAG). As shown in Figure~\ref{fig:cross},  
replacing the reranker with our reranker while keeping other generators (Self-RAG and InstructRAG) fixed yields only marginal improvements and occasionally leads to performance degradation. This suggests that the standalone effectiveness of our reranker is relatively limited.
However, when paired with our generator, CoRAG consistently outperforms the variant that combines our generator with BGE-Reranker. This contrast reveals an inherent tension between reranking effectiveness and generator sensitivity: jointly optimizing the reranker and generator encourages the generator to rely less on fine-grained ranking signals, which in turn limits the observable gains from further improving the reranker in isolation. Nevertheless, the strong performance of CoRAG indicates that its reranker and generator are well aligned and mutually reinforcing when optimized together.



\begin{table}[t]
\centering
\caption{Ablation study.}
\small
\begin{tabular}{lcccc}
  \hline
  & PopQA & NQ & TriviaQA & 2Wiki \\
  \hline
  RTrain & 66.19 & 62.71 & 75.57 & 49.36 \\
  GTrain & 66.54 & 63.37 & 76.25 & 51.53 \\
  RGReplace & 65.83 & 63.21 & 74.97 & 49.72 \\
  GReplace & 66.26 & 62.46 & 75.00 & 50.00 \\
  RReplace & 70.12 & 69.77 & 80.76 & 56.05 \\
  CoRAG (Ours)& \textbf{71.26} & \textbf{72.49} & \textbf{81.00} & \textbf{58.26} \\
  \hline
\end{tabular}
\label{tab:ablation}
\end{table}


\begin{figure}[t]
    \centering
    \includegraphics[width=0.5\textwidth]{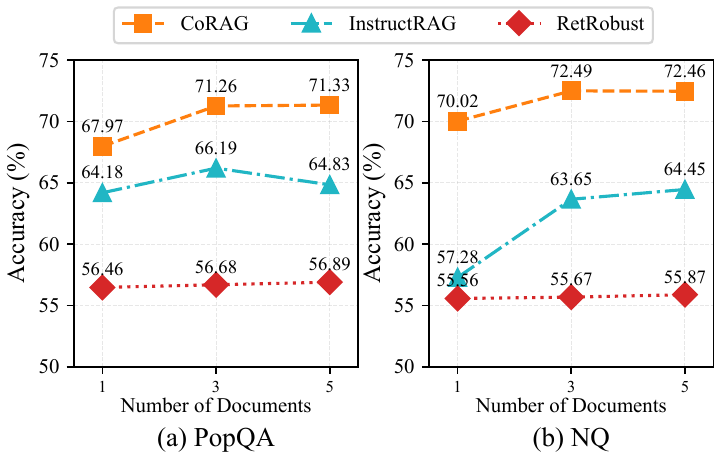}  
    \caption{Impact of the document number.} 
    \label{fig:count} 
\end{figure}

\begin{table}[t]
\centering
\caption{The cross-task evaluation. The InstructRAG in the table is trained on the Pop dataset. WTQ represents WikiTable Questions. The best and second-best performances are highlighted in bold and with an underline. }
\small
\tabcolsep=4pt 
\begin{tabular}{lccc}
\hline
Metric & Llama-3-Instruct\(_{8\text{B}}\) & InstructRAG & CoRAG \\ 
\hline
\multicolumn{4}{l}{\textbf{human-eval}} \\
pass@1 & \textbf{64.45} & 55.85 & \underline{63.96} \\
pass@10 & \underline{83.53} & 76.82 & \textbf{85.97} \\
\hline
\multicolumn{4}{l}{\textbf{human-eval+}} \\
pass@1 & \underline{56.58} & 49.26 & \textbf{57.43} \\
pass@10 & \underline{77.43} & 70.12 & \textbf{79.26} \\
\hline
\textbf{WTQ} \\accuracy & 49.10 & \textbf{68.57} & \underline{68.11} \\
\hline
\end{tabular}
\label{tab:generator_comparison}
\end{table}

\subsection{Top-N Analysis  (RQ3)}

To analyze how the number of documents provided to the generator affects performance, we evaluate top-1(T1), top-3(T3), and top-5(T5) settings across different datasets. The results are presented in Figure \ref{fig:count}, which reveal three insights:  (1) CoRAG outperforms InstructRAG and RetRobust under all document count settings on both datasets, this indicates that CoRAG can effectively utilize the effective information of documents under different document number; (2) on PopQA, InstructRAG’s performance declines as document number increases (e.g., dropping from 66.19\% at T3 to 64.83\% at T5), implying more documents may trigger the noise of less relevant documents, and leading to hallucinations. 
In contrast, our CoRAG  demonstrates a consistent upward trend in performance on both PopQA and NQ, showing strong robustness.

\subsection{Cross-Task Evaluation (RQ4)}
To further explore the cross-domain generalization capability of CoRAG, we conduct additional experiments on other tasks, following \cite{wei2024instructrag}. Specifically, we evaluate the generator of our CoRAG and other generator (Llama-3-Instruct\(_{8\text{B}}\) and InstructRAG) on
code generation datasets HumanEval, HumanEval+ \cite{chen2021evaluating}, and table question answering dataset WikiTable Questions \cite{pasupat2015compositional}. Results are summarized in Table \ref{tab:generator_comparison}.

As shown in Table \ref{tab:generator_comparison}, compared with the Llama-3-Instruct\(_{8\text{B}}\) and InstructRAG, 
our generator achieves the best or second-best performance across all tasks: on HumanEval and HumanEval+, CoRAG achieves the strongest results on pass@10 and achieves the best pass@1 performance on the more challenging HumanEval+, indicating improved generation quality and robustness on code generation. On WTQ, a table-based question answering benchmark, the generator of CoRAG achieves competitive accuracy, closely matching the strongest baseline while substantially outperforming Llama-3-Instruct\(_{8\text{B}}\). Overall, these results demonstrate that the generator of CoRAG remains effective across both code generation and table-based reasoning tasks.

\subsection{Evaluation with LLM-as-a-judge (RQ5)}
Following InstructRAG \cite{wei2024instructrag}, We use LLMs as a judgment to further evaluate the generation of our CoRAG. Specifically, for a given question and documents ranked by reranker, responses are generated by generator and its quality is assessed using different LLMs. The specific prompt template is shown in Appendix~\ref{app:Prompt}. We choose LlaMa, GPT, DeepSeek, Qwen as the judge. For LlaMa, we use version Llama-3-Instruct\(_{8\text{B}}\). For GPT, we use version gpt-4o. For DeepSeek, we use version deepseek-v3.2. For Qwen, we use version qwen3-vl-235b-a22b-instruction. The results are presented in Table \ref{tab:pattern_llm} . 
As shown in Table \ref{tab:pattern_llm}, CoRAG consistently outperforms both InstructRAG and RetRobust across all evaluation settings, regardless of the LLM used as the judge. CoRAG achieves the highest scores under every evaluator, including pattern-based metrics as well as LlaMa, GPT, DeepSeek, and Qwen, indicating robust and consistently preferred generation quality. 
Notably, the performance gap is most pronounced under the LlaMa judge, where CoRAG receives higher scores than other LLMs. We attribute this effect to the fact that CoRAG is fine-tuned from LlaMa, which may lead to a higher alignment between CoRAG’s outputs and the LlaMa-based judge.

\begin{table}[t]
\centering
\caption{Evaluation with LLM-as-a-judge (PopQA).}
\small
\begin{tabular}{lccc}
\hline
Method & InstructRAG & RetRobust & CoRAG \\
\hline
Pattern-based & 66.19 & 56.68 & \textbf{71.26} \\
LlaMa &75.63 & 25.52 & \textbf{89.21} \\
GPT & 60.83 & 54.90  & \textbf{65.90} \\
DeepSeek & 59.69 & 34.10 & \textbf{64.26} \\
Qwen & 60.69 & 57.26 & \textbf{66.12} \\
\hline
Average &64.60  & 45.69 & \textbf{71.35} \\
\hline

\end{tabular}
\label{tab:pattern_llm}
\end{table}

\section{Related Work}

RAG enhances LLMs' performance by fusing external documents and is the mainstream solution for knowledge-intensive tasks. Existing research could be categorized into three types based on core efficiency-enhancing mechanisms~\citep{gao2023retrieval, zhao2023survey}. 
The first category is data-driven methods, focusing on the mining and reconstruction of information at the query or document level: Decomposition Prompting \cite{khot2022decomposed}, which splits complex tasks through prompt engineering. EviNoteRAG \cite{dai2025evinote} annotates uncertain information in documents with a note-taking-first approach. HtmlRAG \cite{tan2025htmlrag} uses HTML instead of plain text to preserve semantic structural information.
The second category is model-driven methods, which improve a model’s ability to interpret, filter and utilize retrieved documents through fine-tuning. Representative works include: RetRobust \cite{yoran2023making} enhances robustness through contrastive training with positive and negative samples. InstructRAG \cite{wei2024instructrag} explicitly learns the denoising process through self-synthesized reasoning. DynamicRAG \cite{sun2025dynamicrag}  optimizes the order and quantity of documents used to train the reranker through generator feedback.
The third category is strategy-driven methods, also called agentic RAG, which introduces agentic behaviors to dynamically adjust retrieval and generation strategies.
FLARE \cite{jiang2023active} triggers lookahead retrieval when encountering uncertain tokens during generation. 
SelfRAG \cite{asai2024self} introduces a reflection module to synchronously and dynamically adjust both retrieval and generation.
MA-RAG \cite{nguyen2025ma} decomposes workflows into sub-agents via CoT. ComposeRAG \cite{wu2025composerag} enables modular agent composition. 
MMOA-RAG~\cite{chen2025improving} jointly optimizes multiple RAG modules via MARL, but forces heterogeneous modules with distinct functionalities to share a single LLM, risking inter-module knowledge conflicts and incurring substantial inference overhead.

\section{Conclusion}
In this paper, we reformulate reranking and generation as two equal participants in a cooperative decision-making problem and propose Cooperative Retrieval-Augmented Generation (CoRAG).
Unlike conventional RAG frameworks, where the generator exhibits an asymmetric dependency on the reranker and generation quality is highly sensitive to the reranking results, CoRAG treats the reranker and generator as two peer decision-makers. Specifically, the reranker decides which documents, and in what organization, to present to the generator, while the generator determines how to effectively utilize the provided information to accomplish the generation task. Through this cooperative formulation, CoRAG alleviates the generator’s reliance on overly strict document ordering and enables more robust coordination between retrieval and generation, leading to improved overall performance. 

\section{Limitations}
Although CoRAG improves robustness by reducing the generator’s sensitivity to reranking results, this design may also attenuate the impact of further reranker improvements on generation quality. This reflects an inherent tension between reranking effectiveness and generation sensitivity under joint optimization, which we leave for future exploration.

\section{Acknowledgments}
This work has been supported by the National Key R\&D Program of China (2023YFF0905400), the National Natural Science Foundation of China (Grants No.62307020), and the New Cornerstone Science Foundation through the XPLORER PRIZE.



\bibliography{custom.bib}

\appendix

\section{The training of CoRAG}
\label{appendix_Algorithm1}
The training of CoRAG has been summarized as Algorithm \ref{alg:corag_simplified}.

\begin{algorithm}
\small
\caption{Training of CoRAG}
\label{alg:corag_simplified}
\begin{algorithmic}[1]
    

\FOR{ $t = 1$ to $T$}
    \FOR{ query $q \in Q$}
        \STATE Compute score $s_i = \mathcal{S}_\theta(q, d_i)$;
        \STATE Select top-K documents $D$ according to scores;
        \STATE Generate answer: $\hat{a} = \mathcal{G}_\phi(q, D)$
        \STATE Compute reward: $\mathrm{R}(a^\ast, \hat{a})$;
        \STATE Estimate the expected task success for $d_i \in D$;
        \STATE Sample a stochastic preference label;
        \STATE Update reranker according to $\mathcal{L}_{\text{rank}}$;
        \STATE Update generator according to  $\mathcal{L}_{\text{gen}}$;




    \ENDFOR
\ENDFOR

\end{algorithmic}
\end{algorithm}

\section{Prompt Template}
\label{app:Prompt}
We provide the prompt used for LLM-as-a-Judge in Table~\ref{Prompt of LLM}, and the prompt used by CoRAG in Table \ref{Prompt of Inference}.
\renewcommand{\arraystretch}{1} 
\begin{table}[H]
 \caption{\label{Prompt of LLM}
    Prompt of LLM-as-a-Judge 
  }
\small
\centering
\begin{tabular}{p{7cm}} 
\toprule
\textbf{Prompt}: You are an expert evaluator for large model responses. Your core task is 
to determine whether the large model's response points to any of the correct answers.

Please conduct the evaluation based on the following information:

\hspace*{2em}1. Question: \{question\}

\hspace*{2em}2. List of correct answers: \{answers\}

\hspace*{2em}3. Large model's response: \{response\}

Evaluation Rules:

\hspace*{2em}1. If the core information and key content of the large model's response point to any 
answer in the correct answer list (semantic consistency is acceptable, no need for 
word-for-word matching), please output "yes".

\hspace*{2em}2. If the large model's response deviates from all correct answers, contains obvious 
errors, or fails to effectively respond to the question, please output "no".

\hspace*{2em}3. You only need to output a single word: either "yes" or "no", without any additional 
redundant content.\\
\midrule 
\textbf{Output}: yes\\
\bottomrule
\end{tabular}
\end{table}

\renewcommand{\arraystretch}{1.2} 
\begin{table*}[t]
 \caption{\label{Prompt of Inference}
    Prompt of Inference 
  }
\centering
\small
\begin{tabular}{
  p{13.5cm}
}
\toprule
\textbf{Prompt}: You are tasked with answering the given question by analyzing a set of documents. 
Please follow this STRICT TWO-STEP PROCESS:

    ---

    \textcolor{blue!45!black}{\#\#\# STEP 1: Document Analysis}
    
    For each document:
    
    \hspace*{2em}- First, extract potentially relevant information from the original document. This includes facts, names, dates, or statements that may relate to the question, even if the connection is not immediately obvious.  
    
    \hspace*{2em}- Then, explain the reason for the information extraction in your previous step based on the question. (e.g., how the document addresses the question’s focus).

    \textcolor{blue!45!black}{\#\#\# STEP 2: Final Answer}
    
    \hspace*{2em}- Summarize your answer to the question based on the analysis above.  
    - If none of the documents are helpful or relevant, answer based on your own general knowledge. In that case, clearly state that you are doing so.

    ---

    Use the following format for your response:

    \#\#\# Step 1: Document Analysis  
    
    \hspace*{2em}Document 1:  
    - Extraction: ...  
    - Explanation: ...  

    \hspace*{2em}Document 2:  
    - Extraction: ...  
    - Explanation: ...  

    \#\#\# Step 2: Final Answer  
    
    \hspace*{2em}Well-supported answer, based on the relevant documents. If no relevant documents, answer based on general knowledge and say so explicitly.
    
    ---

\textcolor{orange!80!black}{EXAMPLE:}

    Question: Who is the author of The Mahdi?

    \textcolor{blue!45!black}{\#\#\# Step 1: Document Analysis}
    
    \hspace*{2em}Document 1:  
    - Extraction:  'Mahdi' is a thriller novel by Philip Nicholson written in 1981 under the identity of a. J. Quinnell ...
    - Explanation: This document directly states that the author of 'Mahdi' is Philip Nicholson. Upon re-examining the question "Who is the author of The Mahdi?", the document mentions the corresponding book title and provides the author information requested.

    \hspace*{2em}Document 2:  
    - Extraction: 'The Mahdi' was published by Philip Nicholson in 1981 
under the pen name A.J. Quinnell, establishing his presence in the thriller 
genre with a novel known for its gripping plot and enduring popularity...  
    - Explanation: This document directly states that 'The Mahdi' was 
published by Philip Nicholson. Upon re-examining the question "Who 
is the author of The Mahdi?", the document mentions the corresponding 
book title and provides the author information requested.

    \textcolor{blue!45!black}{\#\#\# Step 2: Final Answer}  
    
    \hspace*{2em}Based on Documents 1 and 2, The answer to question 'Who is the author 
of The Mahdi?' is A.J. Quinnell, a pseudonym of Philip Nicholson.
    
    ---
    
   Now it is your turn to analyze the following documents and answer the 
given question by following the two-step process.

   \textcolor{teal!80!black}{\{context\}    \{question\}} \\

\bottomrule

\end{tabular}
\end{table*}

\clearpage
\onecolumn
\section{Additional Experimental Results}
\label{appendix:C}

C.1 Robustness to Document Order

To validate whether our CoRAG reduces the ranking sensitivity, we conduct shuffle experiments on PopQA and NQ datasets. Specifically, we evaluate Llama-3-Instruct$_{8B}$, the generator of InstructRAG (InstructRAG-Gen), and the generator of CoRAG (CoRAG-Gen), under two conditions: (1) Ranked: documents are provided in the order produced by the reranker; (2) Shuffled: the document order is randomly shuffled before being fed into the generator. All other settings remain identical to the main experiments. 

As shown in Table \ref{tab:shuffle}, while Llama-3-Instruct$_{8B}$ shows a slight 
performance drop on both PopQA and NQ after shuffling, and InstructRAG-Gen suffers 
a more notable drop of 2.96\% on NQ after shuffling, CoRAG-Gen remains stable,
confirming its robustness to document order.

\begin{table}[H]
\centering
\caption{Performance comparison under shuffled vs. ranked document order on PopQA and NQ datasets}
\label{tab:shuffle}
\resizebox{0.5\textwidth}{!}{
\begin{tabular}{lcccc}
  \hline
  & \multicolumn{2}{c}{PopQA} & \multicolumn{2}{c}{NQ} \\
    & Ranked & Shuffle & Ranked & Shuffle\\
  \hline
  Llama-3-Instruct\(_{8\text{B}}\) & 65.83 & 65.68  & 64.65 & 63.68 \\
  \hline
  InstructRAG-Gen & 66.69 & 66.33  & 64.45 & 61.49 \\
  \hline
  CoRAG-Gen & \textbf{71.26} & \textbf{71.05}& \textbf{72.46}  & \textbf{72.10} \\
  \hline
\end{tabular}
}
\end{table}

\vspace{1.2em}

\end{document}